\documentclass[11pt]{article}
\usepackage{syntaxfest2021}
\usepackage{times}
\usepackage{url}
\usepackage{latexsym}
\usepackage[small,bf,skip=5pt]{caption}
\usepackage{subcaption}
\usepackage[colorlinks=true,linkcolor=black,citecolor=black,filecolor=black,urlcolor=black]{hyperref}
\usepackage{nert}
\usepackage{svg}
\usepackage{adjustbox}

\hypersetup{colorlinks=true,linkcolor=black,citecolor=black,filecolor=black,urlcolor=black}

\usepackage{tikz-dependency}
\usepackage{tipa}

\crefname{xnumi}{example}{examples}
\Crefname{xnumi}{Example}{Examples}
\crefname{xnumii}{example}{examples}
\Crefname{xnumii}{Example}{Examples}
\crefformat{xnumi}{(#2#1#3)}

\newcommand{\ud}[1]{{\texttt{#1}}}

\syntaxfestfinalcopy 

\usepackage{leipzig}
\newleipzig{alter}{alter}{alterphoric}
\newleipzig{ppp}{ppp}{past passive participle}
\newleipzig{emph}{emph}{emphatic particle}

\makeglossaries
\usepackage{gb4e}
\noautomath

\title{For the Purpose of Curry: A UD Treebank for Ashokan Prakrit}

\author{Adam Farris$^*$ \\
  San Mateo High School \\
  \eml{adamfarris@gmail.com} \\\And
  Aryaman Arora$^*$ \\
  Georgetown University \\
  \eml{aa2190@georgetown.edu} \\}

\date{}

\begin{document}
\def\thefootnote{$*$}\footnotetext{Equal contribution.}\def\thefootnote{\arabic{footnote}}
\maketitle
\begin{abstract}
We present
the first linguistically annotated treebank of Ashokan Prakrit, an early Middle Indo-Aryan dialect continuum attested through Emperor Ashoka Maurya's 3rd century BCE rock and pillar edicts. For annotation, we used the multilingual Universal Dependencies (UD) formalism, following recent UD work on Sanskrit and other Indo-Aryan languages. We touch on some interesting linguistic features that posed issues in annotation: regnal names and other nominal compounds, ``proto-ergative'' participial constructions, and possible grammaticalizations evidenced by \textit{sandhi} (phonological assimilation across morpheme boundaries).\footnotetext{The example of \textit{sūpātʰāya} `for the purpose of curry' (discussed further in \cref{sec:sandhi}) inspired the title of this paper.

Glossing abbreviations: \printglossaries} Eventually, we plan for a complete annotation of all attested Ashokan texts, towards the larger goals of improving UD coverage of different diachronic stages of Indo-Aryan and studying language change in Indo-Aryan using computational methods.
\end{abstract}

\section{Introduction}

Ashokan Prakrit is the earliest attested stage and among the most conservative known forms of Middle Indo-Aryan (MIA), represented by inscriptions in the form of rock and pillar edicts commissioned by the Mauryan emperor Ashoka (\textit{aśōka}\footnote{Throughout this work, we use a newly devised transliteration scheme,
devised by Samopriya Basu,
based on the International Alphabet of Sanskrit Transliteration (IAST) which is standard in Indological work, as well as influences from the IPA and Americanist systems. Divergences from IAST are: 1. indication of aspiration and breathy voice with superscript $\langle$ʰ$\rangle$, 2. explicit marking of $\langle$ē$\rangle$ and $\langle$ō$\rangle$ as long vowels, 3. overdot for visarga $\langle$ḣ$\rangle$ and anusvara $\langle$ṁ$\rangle$, instead of the underdot, to avoid confusion with retroflexion.}) in the 3rd century BCE. The Indo-Aryan languages are the predominant language family in the northern (and insular southern) parts of the Indian subcontinent, and consitute a branch of the widespread Indo-European family. They are generally divided into three historical stages: Old Indo-Aryan (OIA; Sanskrit, both the language of Vedic and of later Classical texts, as well as unattested varieties suggested by dialectal variation in later stages), Middle Indo-Aryan (MIA; Ashokan Prakrit, Pali, the Dramatic Prakrits, and early koinés of the Hindi Belt), and New Indo-Aryan (NIA; modern Indo-Aryan languages such as Hindi--Urdu, Assamese, Marathi, Dhivehi, Kashmiri, Khowar, etc.).

Diachronically, Ashokan Prakrit is a descendant of Old Indo-Aryan varieties (some of which are attested through Vedic and Classical Sanskrit) and is a precursor to regional fragmentation of Middle Indo-Aryan into Pali, the Dramatic Prakrits, and eventually the NIA languages. Ashokan Prakrit is a dialect continuum rather than a standardized language, but the three dialect zones are not divergent enough to prove mutually unintelligible \cite{oberlies}.

Universal Dependencies \cite{nivre-etal-2016-universal,ud-recent} is a multilingual formalism for treebanking, including annotation guidelines for dependency relations, morphological analysis, part-of-speech tagging, and other linguistic features. Several New Indo-Aryan languages \cite{bhatt-etal-2009-multi,tandon-etal-2016-conversion,ravishankar-2017-universal} and Sanskrit \cite{kulkarni-etal-2020-dependency,hellwig-etal-2020-treebank,dwivedi} have treebanks annotated using UD or other syntactic formalisms, but thus far there is no treebank for a MIA language, leaving a gap in Indo-Aryan historical corpora. Within MIA, Ashokan Prakrit has an unusual corpus of parallel texts representing multiple geographical dialects, conducive to the study of Indo-Aryan linguistic fragmentation using computational tools.

To this end, we began UD annotation of a digitized Ashokan Prakrit corpus under the \textbf{Digitizing Imperial Prakrit Inscriptions} (DIPI)\footnote{From Shahbazgarhi, Mansehra \textit{dipi} `rescript, writing', as opposed to the lateralized variant \textit{lipi} attested in other dialects.} project.
We will present some interesting annotation issues that arose, both in the context of Indo-Aryan comparative linguistics and for the Universal Dependencies annotation scheme, and suggest future directions for historical and dialectological corpus linguistic work in the Indo-Aryan family.

\section{Related work}
\begin{figure}
    \centering{%
    \setlength{\fboxsep}{0pt}%
    \fbox{\adjustbox{viewport=86cm 33cm 98cm 46cm,clip=true,max width=0.5\columnwidth}{\includesvg[width=8\columnwidth]{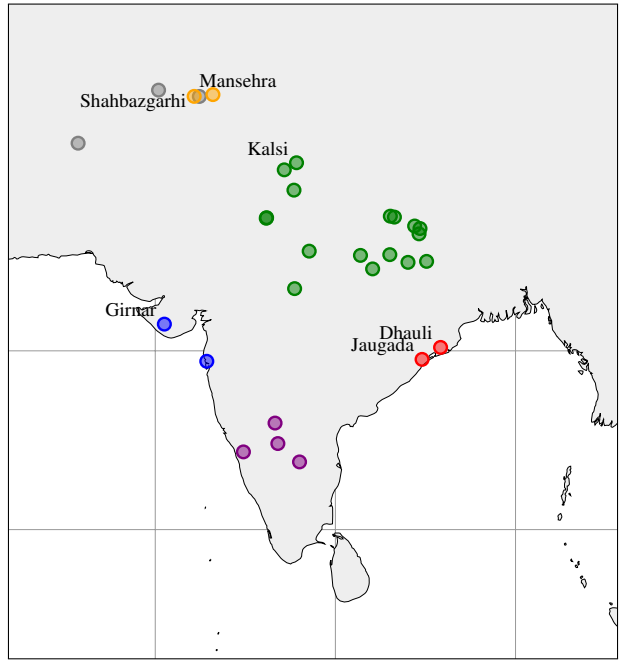}}}%
    }%
    \caption{Locations of the various Ashokan inscriptions and edicts in the Indian Subcontinent, coloured by their usual geographic grouping (not by linguistic isoglosses). Points in grey in the northwest are inscriptions that are not in Ashokan Prakrit (instead, Aramaic and Greek).}
    \label{fig:map}
\end{figure}
The first Ashokan edicts were deciphered by James Prinsep in the 1830s \cite{kopf}. Since then, they have played an important role in the historical study of Ashoka and the Mauryan Empire, sociological and religious study on early Buddhism and other heterodox Dharmic sects \cite{smith2016finding,scott1985ashokan}, and, of course, linguistic work from a historical and social perspective. \Cref{fig:map} shows the locations of the known Ashokan inscriptions, with labels on the locations particularly relevant to this paper.

There are several works which attempt a broad comparative study of the inscriptions with reference to Sanskrit \cite{woolner,hultzsch,mehendale,bloch,sen,oberlies}. Like most historical linguistic work on IA, these works focus mostly on phonology and, to a lesser extent, morphology to the detriment of syntax and semantics \cite{varma}.

On the computational side, the only digitized and machine-readable version of the Ashokan edicts is the Ashoka Library \cite{ashoka-library}, which is sourced from \newcite{hultzsch}
and thus missing more recently discovered inscriptions.

Other UD corpora and their annotation guidelines were also helpful to our own annotation process, e.g.~\newcite{vedicannotation}. Hand-prepared Ashokan Prakrit inflectional tables based on data harvested from \newcite{mehendale} were of use, in addition to Sanskrit dictionaries \cite{mw,epigraphical} and morphological analysers \cite{huet2005functional}.\footnote{The Sanskrit Grammarian \cite{huet2005functional} has a web interface at \url{https://sanskrit.inria.fr/DICO/grammar.html}.}

\section{Corpus}

\begin{figure}[t]
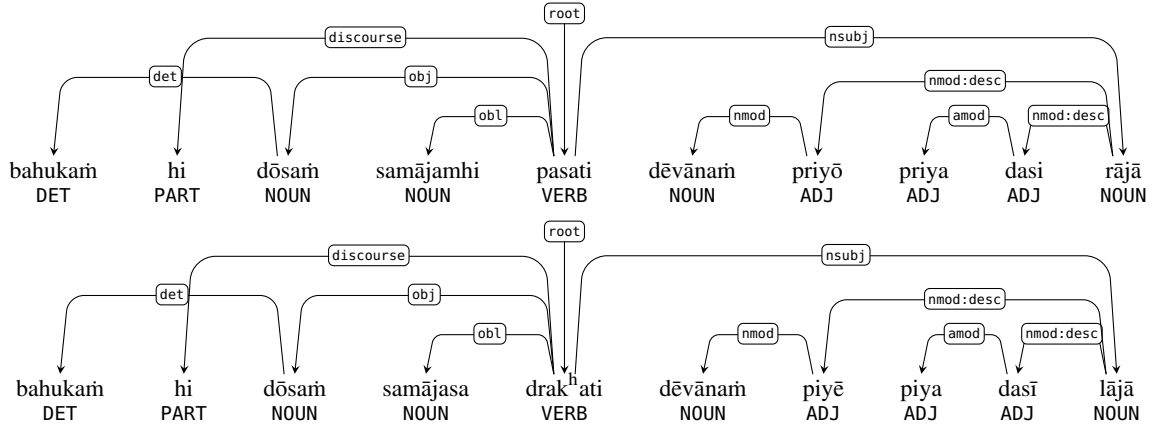

    \small
    \centering
\begin{dependency}
    \begin{deptext}[column sep=0.6cm]
        bahukaṁ \& hi \& d\={o}saṁ \& samājamhi \& pasati \& d\={e}vānaṁ \& priy\={o} \& priya \& dasi \& rājā \\
        \ud{DET} \& \ud{PART} \& \ud{NOUN} \& \ud{NOUN} \& \ud{VERB} \& \ud{NOUN} \& \ud{ADJ} \& \ud{ADJ} \& \ud{ADJ} \& \ud{NOUN} \\
    \end{deptext}
    \deproot{5}{\ud{root}}
    \depedge[edge unit distance=2.2ex]{5}{10}{\ud{nsubj}}
    \depedge{10}{9}{\ud{nmod:desc}}
    \depedge{9}{8}{\ud{amod}}
    \depedge[edge unit distance=2.3ex]{10}{7}{\ud{nmod:desc}}
    \depedge{7}{6}{\ud{nmod}}
    \depedge{5}{4}{\ud{obl}}
    \depedge{5}{3}{\ud{obj}}
    \depedge{3}{1}{\ud{det}}
    \depedge{5}{2}{\ud{discourse}}
\end{dependency}
\begin{dependency}
    \begin{deptext}[column sep=0.6cm]
        bahukaṁ \& hi \& d\={o}saṁ \& samājasa \& drak\textsuperscript{h}ati \& d\={e}vānaṁ \& piy\={e} \& piya \& dasī \& l\={a}j\={a} \\
        \ud{DET} \& \ud{PART} \& \ud{NOUN} \& \ud{NOUN} \& \ud{VERB} \& \ud{NOUN} \& \ud{ADJ} \& \ud{ADJ} \& \ud{ADJ} \& \ud{NOUN} \\
    \end{deptext}
    \deproot{5}{\ud{root}}
    \depedge[edge unit distance=2.2ex]{5}{10}{\ud{nsubj}}
    \depedge{10}{9}{\ud{nmod:desc}}
    \depedge{9}{8}{\ud{amod}}
    \depedge[edge unit distance=2.3ex]{10}{7}{\ud{nmod:desc}}
    \depedge{7}{6}{\ud{nmod}}
    \depedge{5}{4}{\ud{obl}}
    \depedge{5}{3}{\ud{obj}}
    \depedge{3}{1}{\ud{det}}
    \depedge{5}{2}{\ud{discourse}}
\end{dependency}
    \caption{Dependency parse of the fourth sentence of Major Rock Edict 1 as found in two locations. The top is from Girnar, representing the Western dialect, and the bottom is from Jaugada, representing the Eastern dialect.}
    \label{fig:edict1}
\end{figure}

The Ashokan Prakrit texts available to us constitute a very limited corpus. They are royal inscriptions concerning the promotion of Buddhist morality, administration of the Mauryan Empire, and records of Ashoka's magnanimous deeds (such as his conversion to Buddhism). They directly address the public, and all evidence points to Ashokan Prakrit being a semi-standardized but still fairly accurate reflection of vernacular language, given the geographical dialect variation and communicative purpose of the texts.

We began with transcribed edicts from the Ashoka Library \cite{ashoka-library}. Annotation began in June 2021 and was done in Google Sheets simultaneously by two linguistically-informed annotators with discussions to resolve disagreements. Although Google Sheets is not the conventional choice of tool for such a project, existing UD annotation tools were found to be lacking a convenient means of editing \ud{FEATS} columns in a \textsc{conllu} file, as well as supplying additional columns (e.g. etymologies). Additionally, this allowed us to avoid setting up the server required for collaborative annotation with tools like UD Annotatrix \cite{tyers-etal-2017-UD} A guidelines document was added to as the analysis of tricky constructions was decided upon.

Given the parallel nature of the corpus, annotations for a particular edict at one location could be transferred with little modification to that of another location. An example of this is given in \cref{fig:edict1}, which only shows POS-tag and dependency parse UD annotations of a parallel sentence, glossed below.
\begin{exe}
    \ex \gll bahukaṁ hi d\={o}saṁ samājamhi pasati D\={e}vānaṁ- priy\={o} Priya- dasi rājā \\
        very {\Emph} {evil:\Acc.\M.\Sg} {meeting:\Loc.\M.\Sg} {see:\Prs.\Ind.\Third.\Sg} {god:\Nom.\M.\Pl} {beloved:\Nom.\M.\Sg} {kindly} {looking:\Nom.\M.\Sg} {king:\Nom.\M.\Sg} \\
        \glt `King Beloved-of-the-Gods Looking-Kindly sees much evil in festival meetings.' \hfill{}(Girnar 1:4)
\end{exe}
Thus, we used the well-preserved edicts at Girnar as the main annotation document, and annotated other editions only after finalising the corresponding Girnar version. \Cref{tab:stats} gives statistics about the annotated corpus.

\section{Annotation and analysis}

\begin{table}[t]
    \centering
    \begin{subtable}[t]{0.49\textwidth}
    \centering
        \begin{tabular}[t]{lrrr}
        \toprule
        & \textbf{Doc.} & \textbf{Sent.} & \textbf{Tok.} \\
        \midrule
        Girnar & 5 & 43 & 534 \\
        Shahbazgarhi & 3 & 14 & 158 \\
        Mansehra & 1 & 8 & 87 \\
        Kalsi & 1 & 8 & 85 \\
        Jaugada & 1 & 8 & 89 \\
        Dhauli & 1 & 3 & 20 \\
        \midrule
        \textbf{Total} & 12 & 84 & 973 \\
        \bottomrule
        \end{tabular}
        \caption{DIPI corpus composition, grouped by source location of the annotated inscriptions.}
        \label{tab:stats}
    \end{subtable}
    \begin{subtable}[t]{0.49\textwidth}
    \centering
        \begin{tabular}[t]{llr}
        \toprule
        \textbf{Feature} & \textbf{Measure} & \textbf{Val.} \\
        \midrule
        \ud{UPOS} & Cohen's $\kappa$ & 0.949 \\
        \ud{HEAD} & UAS & 0.857 \\
        \ud{DEPREL} & Label score & 0.776 \\
        \ud{HEAD}+\ud{DEPREL} & LAS & 0.673 \\
        \bottomrule
        \end{tabular}
        \caption{Agreement scores between two annotators on Girnar Major Rock Edict 7.}
        \label{tab:agreement}
    \end{subtable}
    \caption{Metrics about the DIPI corpus.}
\end{table}

We annotated using the standard Universal POS tag inventory and Universal Dependency Relations from Universal Dependencies v2, with some additional dependency subtypes: \ud{acl:relcl}, \ud{advmod:lmod}, \ud{advmod:tmod}, \ud{advmod:neg}, \ud{nmod:desc} (discussed in \cref{sec:titles}), \ud{obl:lmod}, \ud{obl:tmod}. Overall UPOS counts are given in \cref{tab:upos}.

Most of the corpus was annotated collaboratively with continuous revisions to maximize annotation quality given the lack of reliable modern grammars and lexicons for Ashokan Prakrit. Major Rock Edict 7 at Girnar (5 sentences, 49 tokens) was annotated by both authors independently to compute interannotator agreement figures. Agreement scores are reported in \cref{tab:agreement}. Agreement on universal POS tagging and head attachment is high. Low labelled attachment score (LAS) reflects the difficulty in analysing the sometimes fragmentary language of the corpus, as is expected in treebanking ancient language corpora \cite{greek}.

The most common (and thus likely pragmatically unmarked word order, modulo the inscriptional nature of the corpus) in Ashokan Prakrit is subject--object--verb, occuring in half of 24 verbs in the corpus with \ud{nsubj} and \ud{obj} dependents, followed by object--subject--verb with 8 occurrences. SOV is the unmarked word order in most New Indo-Aryan languages as well.

\begin{table}[t]
    \centering
    \begin{subtable}[t]{0.3\textwidth}
    \centering
        \begin{tabular}{lrr}
        \toprule
        \textbf{UPOS} & \textbf{Count} & \textbf{\%} \\
        \midrule
        \ud{NOUN} & 345 & 35.5\% \\
        \ud{ADJ} & 136 & 14.0\%\\
        \ud{VERB} & 106 & 10.9\%\\
        \ud{ADV} & 83 & 8.5\%\\
        \ud{CCONJ} & 78 & 8.0\%\\
        \ud{PRON} & 47 & 4.8\%\\
        \ud{PART} & 42 & 4.3\%\\
        \bottomrule
        \end{tabular}
    \end{subtable}
    \begin{subtable}[t]{0.3\textwidth}
        \centering
        \begin{tabular}{lrr}
        \toprule
        \textbf{UPOS} & \textbf{Count} & \textbf{\%} \\
        \midrule
        \ud{DET} & 42 & 4.3\%\\
        \ud{NUM} & 35 & 3.6\% \\
        \ud{PROPN} & 22 & 2.3\%\\
        \ud{X} & 14 & 1.4\%\\
        \ud{SCONJ} & 11 & 1.1\%\\
        \ud{ADP} & 10 & 1.0\%\\
        \ud{\_} & 2 & 0.2\%\\
        \bottomrule
        \end{tabular}
    \end{subtable}
    \caption{Top UPOS categories. \ud{PUNCT}, \ud{SYM}, and \ud{INTJ} were not used.}
    \label{tab:upos}
\end{table}

\section{Annotation issues}

Some of the interesting annotation issues faced include: the POS-tagging and dependency parsing of regnal names in Ashokan Prakrit and cross-lingually (with further discussion on compounds in general), the in-progress transition to split ergativity and its morphological and syntactic analysis within the framework of UD, as well as the relationship between irregular sandhi and the grammaticalization of nouns into adpositions.

A recurring point in the analysis of these issues is that Ashokan Prakrit is transitional between Sanskrit and New Indo-Aryan, still in the process of undergoing many drastic syntactic (from non-configurational to configurational) and morphological (from synthetic to analytic) changes. Given the small size of the corpus and inability to elicit information from native speakers, we faced difficulties annotating features based on a synchronic analysis without looking towards better, and often conflicting, data from Sanskrit or NIA languages.

\subsection{Regnal names}

A puzzling issue in annotation was Ashoka's regnal names, such as:
\begin{exe}
    \ex \gll Dēvānaṁ- priyēna Priya- dasinā rāña\\
        {god:\Gen.\M.\Pl} {beloved:\Ins.\M.\Sg} {kindly} {looking:\Ins.\M.\Sg} {king:\Ins.\M.\Sg} \\
    \glt `King Beloved-of-the-Gods Looking-Kindly' \hfill (Girnar 1:1)
\end{exe}
Ashokan Prakrit, like Sanskrit, often constructs chains of nominals and adjectives headed by the last member and with all the members agreeing in case and number with it---here, the instrumental singular. Tokenization, POS-tagging of morphemes in compounds, and dependency relations in regnal names all came up as issues in UD annotation. The decisions in this section were arrived at after much discussion with the UD community.\footnote{Documented in a GitHub issue: \url{https://github.com/UniversalDependencies/docs/issues/802}.}

\subsubsection{POS annotation of morphemes in compounds}\label{sec:titles}

The first issue was how to POS tag the morphemes in such compounds. In Ashokan Prakrit, like in Sanskrit, ``the division-line between substantive and adjective ... [is] wavering'' \cite{whitney} so any of these titles could be thought of as nominals (`one who is beloved of the Gods') or adjectives (`beloved by the Gods'). Furthermore, syntactic context can blur the distinction; an adjective like \textit{dasin} `looking' can be nominalized into `looker', and a noun in a compound may behave attributively.

Initially, we thought to label all the morphemes in the regnal names as \ud{PROPN} given that they refer to a person like a regular name does. However, these morphemes have internal dependency structure, most obviously the genitive-case modifier in \textit{dēvānaṁ-priyēna}. The \ud{PROPN} label would obscure what is clearly a genitive-case \ud{NOUN}, \textit{dēvānaṁ} `of the Gods', that does not refer to a specific individual or entity like a name does.

In regards to differentiating between \ud{NOUN} and \ud{ADJ} in Ashokan Prakrit, we settled on the criterion that something with a fixed inherent gender must be \ud{NOUN}, and anything with fluid gender assignment is \ud{ADJ}. This makes the POS tag a lexical feature rather than one that is contextually assigned by syntactic properties, which would render it redundant. UD precedent in other languages, e.g.~Latin, favours the annotation of dependency structure in proper nouns and the regular POS tagging of nominalized components in such names.\footnote{\url{https://github.com/UniversalDependencies/docs/issues/777}}

\subsubsection{Dependency structure of nominalized titles}

\begin{figure}
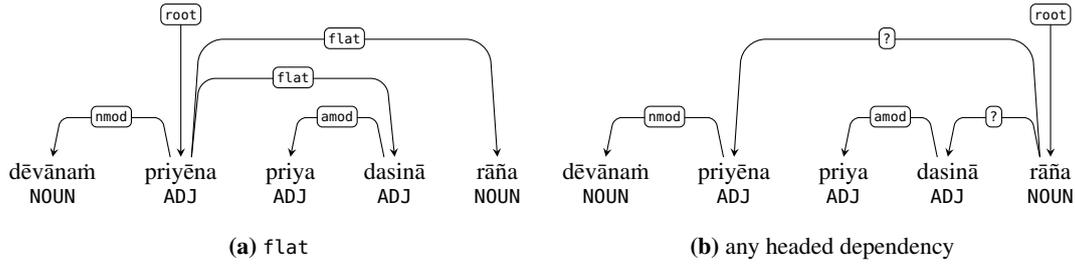

    \small
    \centering
\begin{subfigure}[b]{0.45\textwidth}
\begin{dependency}
    \begin{deptext}[column sep=0.5cm]
        dēvānaṁ \& priyēna \& priya \& dasinā \& rāña \\
        \ud{NOUN} \& \ud{ADJ} \& \ud{ADJ} \& \ud{ADJ} \& \ud{NOUN} \\
    \end{deptext}
    \deproot{2}{\ud{root}}
    \depedge{2}{4}{\ud{flat}}
    \depedge{4}{3}{\ud{amod}}
    \depedge{2}{5}{\ud{flat}}
    \depedge{2}{1}{\ud{nmod}}
\end{dependency}
\caption{\ud{flat}}
\end{subfigure}
\begin{subfigure}[b]{0.45\textwidth}
\begin{dependency}
    \begin{deptext}[column sep=0.5cm]
        dēvānaṁ \& priyēna \& priya \& dasinā \& rāña \\
        \ud{NOUN} \& \ud{ADJ} \& \ud{ADJ} \& \ud{ADJ} \& \ud{NOUN} \\
    \end{deptext}
    \deproot{5}{\ud{root}}
    \depedge{5}{4}{\ud{?}}
    \depedge{4}{3}{\ud{amod}}
    \depedge{5}{2}{\ud{?}}
    \depedge{2}{1}{\ud{nmod}}
\end{dependency}
\caption{any headed dependency}
\end{subfigure}
    \caption{Potential dependency parses (headless and headed) of Ashoka's regnal names.}
    \label{fig:regnal}
\end{figure}

There is substantial disagreement among UD corpora on the dependency annotation of regnal names, epithets, and other appellative titles. The current UD guidelines prefer the \ud{flat} relation for ``exocentric (headless) semi-fixed MWEs [multi-word expressions] like names and dates''. The head is arbitrarily assigned to be the first nominal in the multi-word expression. This is unacceptable for titles in Ashokan Prakrit, since want to treat this the same way as adjective--noun NPs, with the head always being the last word. \newcite{schneider2021mischievous} recently attempted to resolve this issue for a wide range of nominal constructions in English (including \textit{Mr.} and \textit{Secretary of State}, which are similar to Ashokan Prakrit titles), and we build upon that analysis here.

Since we have established that in Ashokan Prakrit such constructions are not headless, we have to decide which headed dependency relation should be used instead. We considered \ud{appos}, \ud{compound}, and \ud{nmod:desc}, and \ud{amod} if we chose to analyse the appellatives as adjectival rather than nominal. The difference between a headed and headless dependency analysis of the regnal titles is shown in \cref{fig:regnal}.

The issues, resolved once we came to \ud{nmod} after settling our POS tagging, in the other relations are:
\begin{itemize}
    \item \ud{appos}: Generally, an appositive is a full NP that can be paraphrased with an equational copula in a relative clause, e.g. \textit{Bob, my friend} implies \textit{Bob, who is my friend}. But in Ashokan Prakrit, given the blurring between nouns and adjectives, it is clear that each title NP is directly modifying the NP \textit{rāña} `king' rather than paraphrasing an appositional relationship.
    \item \ud{compound}: Like \ud{flat}, this indicates a multiword expression forming a single NP rather than relationships between full NPs. Each regnal name is, however, a whole NP that could stand alone.
    \item \ud{amod}: Our reasoning against the other two relies on analysing each title as an NP. The fact that titles can be dropped, and that \textit{rāña} `king' can be dropped while retaining grammaticality, supports the assumption that each title is indeed an NP since any one could be the head if phrase-final. Thus, an adjectival relation like \ud{amod} is not preferred.
\end{itemize}
Realising that the head of each NP title is lexically a nominalized \ud{ADJ}, we settled on \ud{nmod:desc} as the best dependency relation. Further evidence comes from variation in the components of the titles in different editions of the edicts, e.g.~\cref{ex:drop-kind} and \cref{ex:drop-king}. Given that \cref{ex:drop-king} drops `king' entirely and can have the titles stand alone without another NP head, we are certain that each title is an NP.
\begin{exe}
    \ex \label{ex:drop-kind}\gll Dēvana- priasa rañ\={o}\\
             {god:\Gen.\M.\Pl} {beloved:\Gen.\M.\Sg} {king:\Gen.\M.\Sg} \\
        \glt `King Beloved-of-the-Gods' \hfill (Shahbazgarhi 1:1)
    \ex \label{ex:drop-king}\gll Dēvānaṁ- piyēna Piya- das[i]nā\\
             {god:\Gen.\M.\Pl} {beloved:\Ins.\M.\Sg} {kindly} {looking:\Ins.\M.\Sg} \\
        \glt `Beloved-of-the-Gods Looking-Kindly' \hfill (Kalsi 1:1)
\end{exe}
Now backed with our crosslinguistic evidence, we agree with \newcite{schneider2021mischievous} that \ud{nmod} or a subtyped label of it is the best descriptor for nominal epithets. We specifically picked the subtyped label so that we can query instances of the construction for future analysis.

\subsection{Predicated \textit{-ta} construction}

The \textit{-ta} construction\footnote{Philologically known as the \textit{past passive participle}.} in Sanskrit forms participles from verbal roots. These participles are morphologically deverbal adjectives, taking gender (without having intrinsic fixed gender like nouns), case, and number marking without marking person (unlike finite verb forms).
\begin{exe}
    \ex \gll rājñā \textbf{hataḣ} cauraḣ\\
        {king:\Ins.\M.\Sg} {kill:\Ppp.\Nom.\M.\Sg} {thief:\Nom.\M.\Sg}\\
        \glt `a thief killed by a king' (lit.~`a king-killed thief')\hfill{}(Sanskrit)
\end{exe}
In Sanskrit, especially in post-Vedic texts, it can also be interpreted with (past) perfect meaning. -\textit{ta} forms agree in case/gender/number with the object, unlike the finite verbs of this stage of Indo-Aryan.
\begin{exe}
    \ex \gll mayā lipī \textbf{likʰitā}\\
        {\First\Sg.\Ins} {text:\Nom.\F.\Sg} {write:\Ppp.\Nom.\F.\Sg}\\
        \glt `the text was written by me' (passive)\\
            `I wrote the text' (ergative)\hfill{}(Sanskrit)
\end{exe}

This use is extremely common in Ashokan Prakrit and is the point of contention discussed here. According to one view, \textit{-ta} formed resultative\footnote{As opposed to stative adjectives, resultatives imply that a prior event occurred to cause the current state conveyed by the adjective. Compare English \textit{is hidden} with \textit{has been hidden} \cite{condoravdi20149}.} adjectives in early OIA, gradually shifting towards main predicate function in first intransitive and later transitive verbs (the agent receiving case marking) by late OIA \cite{reinohl2018,condoravdi20149,peterson1998}.

This construction is ancestral to the tense/aspect-based split ergativity observed in many later NIA languages. In such languages, the Sanskrit participle has developed into a perfect verb that agrees with the object, while other inflected forms in the verb paradigm agree with the subject. Since Ashokan Prakrit was still undergoing this transition to split ergativity, we could analyze this construction either way: as a resultative predicate adjective or a perfect-aspect verb.

In Ashokan Prakrit, with the loss of the inherited active aorist as a productive category, \textit{-ta} forms have become the unmarked strategy to express the past perfect \cite{bubenik1998}. We believe, with some certainty, that this construction is \textit{not passive} at least as late as Ashokan Prakrit (if it ever was). Evidence \newcite{casaretto2020} provide against a passive analysis in Sanskrit also applies here. A key argument is that \textit{-ta} occurs with both transitive and intransitive verbs, and in the case of the latter, does not form an ``impersonal passive'' as would be expected of a passivized intransitive. 

As such, we adopt an ergative-like analysis of the \textit{-ta} construction in Ashokan Prakrit, agreeing with \newcite{peterson1998}'s view of the corresponding construction in Pali (another early MIA lect) as being a periphrastic perfect. Indeed, as exhibited in the example in \cref{fig:ergative} which is glossed in \cref{ex:mansehra}, the \textit{-ta} form agrees in number and gender with the object, while the agent receives instrumental marking. The object \textit{dʰrama-dipi} is still in the nominative case.

\begin{exe}
    \ex \label{ex:mansehra}\gll ayi dʰrama- dipi D\={e}vanaṁ- priy\={e}na Priya- dra\'{s}ina rajina likʰapita \\
        {\Dem\Third:\F.\Sg} {morality} {rescript:\Nom.\F.\Sg} {god:\Nom.\M.\Pl} {beloved:\Ins.\M.\Sg} {kindly} {looking:\Ins.\M.\Sg} {king:\Ins.\M.\Sg} {write:\Caus.\Ppp.\Nom.\F.\Sg} \\
        \glt `King Beloved-of-the-Gods Looking-Kindly has caused this rescript on morality to be written'\hfill(Mansehra 1:1)
\end{exe}

With respect to UD annotation, our ergative-like analysis translates to the agent \textit{rajina} receiving the \ud{DEPREL} \ud{nsubj} and the object \textit{dipi} \ud{obj} (instead of \ud{obl:agent} and \ud{nsubj:pass} of the passive analysis in \cref{fig:passive}).

\begin{figure}
    \small
    \centering
\begin{subfigure}[b]{\textwidth}
    \centering
\begin{dependency}
     \begin{deptext}[column sep=0.6cm]
        ayi \& dʰrama \& dipi \& D\={e}vanaṁ- \& priy\={e}na \& Priya- \& dra\'{s}ina \& rajina \& likʰapita \\
        \ud{PRON} \& \ud{NOUN} \& \ud{NOUN} \& \ud{NOUN} \& \ud{ADJ} \& \ud{ADJ} \& \ud{ADJ} \& \ud{NOUN} \& \ud{VERB} \\
    \end{deptext}
    \deproot{9}{\ud{root}}
    \depedge{9}{8}{\ud{nsubj}}
    \depedge{3}{1}{\ud{det}}
    \depedge{3}{2}{\ud{compound}}
    \depedge[edge unit distance=2.2ex]{9}{3}{\ud{obj}}
    \depedge{5}{4}{\ud{nmod}}
    \depedge[edge unit distance=2.4ex]{8}{5}{\ud{nmod:desc}}
    \depedge{7}{6}{\ud{amod}}
    \depedge{8}{7}{\ud{nmod:desc}}
\end{dependency}
\caption{The ergative-like analysis, with \ud{nsubj} and \ud{obj}\label{fig:ergative}}
\end{subfigure}
\begin{subfigure}[b]{\textwidth}
\centering
\begin{dependency}
    \begin{deptext}[column sep=0.6cm]
        ayi \& dʰrama \& dipi \& D\={e}vanaṁ- \& priy\={e}na \& Priya- \& dra\'{s}ina \& rajina \& likʰapita \\
        \ud{PRON} \& \ud{NOUN} \& \ud{NOUN} \& \ud{NOUN} \& \ud{ADJ} \& \ud{ADJ} \& \ud{ADJ} \& \ud{NOUN} \& \ud{VERB} \\
    \end{deptext}
    \deproot{9}{\ud{root}}
    \depedge{9}{8}{\ud{obl:agent}}
    \depedge{3}{1}{\ud{det}}
    \depedge{3}{2}{\ud{compound}}
    \depedge[edge unit distance=2.2ex]{9}{3}{\ud{nsubj:pass}}
    \depedge{5}{4}{\ud{nmod}}
    \depedge[edge unit distance=2.4ex]{8}{5}{\ud{nmod:desc}}
    \depedge{7}{6}{\ud{amod}}
    \depedge{8}{7}{\ud{nmod:desc}}
\end{dependency}
\caption{The passive analysis, with \ud{obl:agent} and \ud{nsubj:pass}\label{fig:passive}}
\end{subfigure}
\caption{Two possible analyses of the predicated \textit{-ta} construction in the sentence '\textit{king Beloved-of-the-Gods Looking-Kindly has caused this rescript on morality to be written}' (Mansehra 1:1). The above was ultimately chosen.\label{fig:ergativeanalysis}}
\end{figure}

\subsubsection{Differential agent marking}

Cross-dialectally as well as dialect-internally, Ashokan Prakrit varies with respect to how the agent phrase is marked in \textit{-ta} constructions. Agents may receive either \textbf{instrumental} or (with lesser frequency) \textbf{genitive} case marking, though the basis for this alternation is not wholly clear.
\begin{exe}
    \ex \gll s\={e} \textbf{mamayā} bahu kayān\={e} ka\d{t}\={e}\\
             {now} {1\Sg.\textbf{\Ins}} {many} {good\_deed:\Nom.\N.\Sg} {do:\Ppp.\Nom.\N.\Sg} \\
        \glt `Now, I did many good deeds.' \hfill (Kalsi 5:4)
    \ex \gll D\={e}v\={a}naṁ- piy\textbf{a\'{s}a} Piya- da\'{s}\textbf{in\={e}} l\={a}j\textbf{in\={e}} Kaligy\={a} vijit\={a} \\
         {god:\Nom.\M.\Pl} {beloved:\textbf{\Gen}.\M.\Sg} {kindly} {looking:\textbf{\Gen}.\M.\Sg} {king:\textbf{\Gen}.\M.\Sg} {Kalinga:\Nom.\M.\Pl} {conquer:\Ppp.\Nom.\M.\Pl} \\
        \glt `... king Beloved-of-the-Gods Looking-Kindly conquered the Kalingas.' \hfill (Kalsi 13:1)
\end{exe}
\newcite{andersen1986}'s analysis suggests discourse-pragmatic factors may be at play; the genitive agent conveys old (i.e. contextually given and/or definite) information while the instrumental agent conveys new information. On this basis, he also claims these represent two \textit{separate} constructions, a passive and an ergative respectively, though this proposal has some flaws (see \cite{bubenik1998} for criticisms). 

We tentatively follow \newcite{dahl-and-stronski2016} in analyzing the situation as one of \textbf{differential agent marking (DAM)} \cite{arkadiev2017}, whereby two agent-marking cases are distributed along (potentially irrecoverable) semantic/pragmatic lines. Thus we stuck with standard morphological analysis of the case features in these agents, i.e.~\ud{Case=Gen/Ins} rather than explicitly proposing \ud{Case=Erg} as an Ashokan Prakrit feature.

DAM seems to affect both the agents of the ergative-like \textit{-ta} construction as well as the oblique agents of finite passives in Ashokan Prakrit. Of the source constructions in Vedic, \newcite{bubenik1998} explains there is a broad tendency for `active' verbs to favor instrumental agents, and `ingestive' verbs (perception, consumption, etc.) to favor the genitive, but the instrumental becomes default in later stages of Old Indo-Aryan. Further annotation of the Ashokan Prakrit corpus will allow us to probe into these hypotheses with statistical tools.

Additionally, the influence of Ashoka's administrative language, an eastern dialect from which other dialectal edicts were likely translated \cite{oberlies}, should not be neglected. If the choice between instrumental and genitive marking is at least partially a function of dialect, direct translation from Ashoka's variety could leave relic forms\footnote{One such example of dialectal interference is \Nom.\M.\Sg \ \textit{-\={e}}, a non-western isogloss, attested in place of the expected -\={o} in Girnar (a western dialect) 
}  (otherwise inconsistent with the internal distribution of cases) in other edicts.  

\subsection{Sandhi}\label{sec:sandhi}

Sanskrit texts (which in written form all post-date the Ashokan edicts) generally orthographically indicate \textit{sandhi}, a kind of phonological assimilation at morpheme boundaries \cite{sandhi}. Some examples from Sanskrit are given in \cref{ex:sandhi}.

\begin{exe}
    \ex \label{ex:sandhi}\begin{xlist}
        \ex gaccʰat\textbf{i} arjunaḣ $\to$ gaccʰat\textbf{y}arjunaḣ\hfill{}(Sanskrit)
        \ex s\textbf{aḣ a}ham $\to$ s\textbf{\={o}'}ham
        \ex brahm\textbf{a a}smi $\to$ brahm\textbf{ā}smi
    \end{xlist}
\end{exe}
Middle Indo-Aryan has more haphazard orthographic indication of sandhi rules \cite{dockalova2009}, even though these assimilations likely persisted in speech. For example, Pali shows sandhi in compounds (especially those inherited directly from Old Indo-Aryan and then subject to normal sound changes), some function words (emphatic \textit{\={e}va}, preverbs, etc.), pronouns, and sometimes in nominal arguments to verbs, noun--noun relations, and vocatives \cite{childers}. That is, Pali optionally indicates sandhi only between syntactically related words \cite[p.~116]{oberlies2}.

We observed similar occurrences in the Ashokan Prakrit corpus. We think certain rare cases of sandhi in Ashokan Prakrit may be examples of grammaticalization (the development of a postposition with case-like properties) and lexicalization (compounds that are no longer as transparent). These pose issues for UD annotation.

\subsubsection{Grammaticalization of \textit{atʰāya} \textasciitilde{} \textit{aṭʰāya}}

One case where sandhi may gives us clues about morphological change is occurrences of \textit{atʰāya} `for the purpose [of]', the dative of \textit{atʰa} `purpose' (< Sanskrit \textit{ártʰa}). In the prototypical example below, sandhi with the preceding nominal stem causes vowel lengthening.
\begin{exe}
    \ex \gll tī \={e}va prāṇā ārabʰar\={e} sūp- \textbf{ātʰāya}\\
            three {\Emph} {animal:\Nom.\N.\Pl} {kill:\Pass.\Prs.\Ind.\Third\Pl} curry {purpose:\Dat.\M.\Sg} \\
        \glt `Only three animals are being killed for the purpose of curry.'\hfill{}(Girnar 1:7)
\end{exe}
While the Ashokan Prakrit construction we are dealing with is a compound\footnote{
A similar construction involving ``compounded'' \textit{artʰāya} also occurs in certain Sanskrit texts, cf. \textit{har\d{s}a\d{n}\={a}rt\textsuperscript{h}\={a}ya} \textpipe\textit{har\d{s}a\d{n}a + art\textsuperscript{h}\={a}ya}\textpipe \ `for the purpose of protection'\cite{fritz}.
}, not a genitive noun modifier, \newcite{reinohl} describes a potentially related phenomenon based on Classical Sanskrit and Pali corpora: the \textbf{post-Vedic genitive shift}, wherein many adverbs and adjectives were analysed as taking the genitive and periphrastically replacing case relations, e.g. \textit{-asya artʰāya} `for the purpose of ...'. Sanskrit generally uses the dative case by itself to indicate \textsc{Purpose}, so this compounded construction in Ashokan Prakrit may be an intermediate phase in the genitive shift.

\begin{figure}
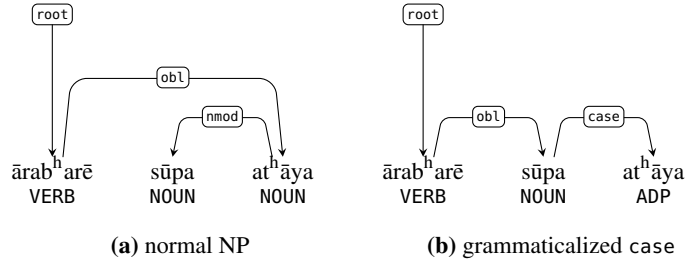

    \small
    \centering
\begin{subfigure}[b]{0.3\textwidth}
\begin{dependency}
    \begin{deptext}[column sep=0.6cm]
        ārabʰar\={e} \& sūpa \& atʰāya \\
        \ud{VERB} \& \ud{NOUN} \& \ud{NOUN} \\
    \end{deptext}
    \deproot{1}{\ud{root}}
    \depedge{3}{2}{\ud{nmod}}
    \depedge{1}{3}{\ud{obl}}
\end{dependency}
\caption{normal NP}
\end{subfigure}
\begin{subfigure}[b]{0.3\textwidth}
\begin{dependency}
    \begin{deptext}[column sep=0.6cm]
        ārabʰar\={e} \& sūpa \& atʰāya \\
        \ud{VERB} \& \ud{NOUN} \& \ud{ADP} \\
    \end{deptext}
    \deproot{1}{\ud{root}}
    \depedge{1}{2}{\ud{obl}}
    \depedge{2}{3}{\ud{case}}
\end{dependency}
\caption{grammaticalized \ud{case}}
\end{subfigure}
    \caption{Two potential analyses of the \textit{atʰāya} construction in Girnar 1:7.}
    \label{fig:purpose}
\end{figure}

For UD, this is a tricky situation. We were stuck between describing \textit{atʰāya} as a \ud{case} complement to \textit{sūpa}, or as instead the head of an NP, both shown in \cref{fig:purpose}. Given similar constructions with ending-less nouns in compounds, \textit{atʰāya} would usually be analysed as the head here, but if it has been grammaticalized then \ud{case} would be a better \ud{DEPREL} as is used for case markers and clitics in New Indo-Aryan UD, and UD prefers content heads. Girnar 4:10 also has \textit{\={e}tāya atʰāya} `for this purpose' which lacks sandhi or stem-compounding, but this may be exceptional since \textit{\={e}tad} can take the \ud{det} \ud{DEPREL} as a modifier to nouns and so does not behave like a true nominal. Pending better evidence supporting either analysis, we settled on the latter.

An interesting data point is that a similar construction is the etymological source of the dative case in the Insular Indo-Aryan languages, Dhivehi and Sinhala. 

\begin{exe}
    \ex \gll mamma e=ge-\textbf{a\d{\r{s}}} diya \\
        {mother} {\Dem3=house-\textbf{\Dat} } {go.\Pst.\Alter} \\
    \glt `Mother went to that house.' (adapted from Lum, \shortcite{lum2020}: 118) \hfill{}(Dhivehi)
\end{exe}

\begin{exe}
    \ex \gll amm\={a} \={e} ged\textschwa r\textschwa-\textbf{\d{t}\textschwa{}} giy\={a} \\
            {mother} {\Dem4} {house.\Sg-\textbf{\Dat}} {go.\Pst} \\
        \glt `Mother went to that house.' \hfill{}(Sinhala)
\end{exe}

The Sinhala -\textit{(\textschwa)\d{t}\textschwa} and Dhivehi \textit{-a\d{\r{s}}} datives are both reflexes of Sanskrit \textit{ártʰāya} (or, possibly, the accusative case \textit{ártʰaṁ}) \cite{fritz} and have expanded their semantic domains to include other roles such as \textsc{goal}. Ashokan Prakrit's compounding of \textit{atʰāya} may represent an early stage towards a similar grammaticalization, though its precise synchronic status is unclear. Future UD annotation of MIA corpora will allow us to better track such phenomena from a comparative perspective.\footnote{It is worth noting that an inscriptionally-attested Middle Indo-Aryan ancestor of Sinhala, roughly contemporaneous with the Ashokan edicts, formed a periphrastic dative of purpose with \textit{a\d{t}aya} (cf. \textit{\'{s}aga\'{s}a a\d{t}aya} `for the benefit of the sangha') \cite{premaratne,paranavithana}. Here, as is also observed with Pali's \textit{attʰāya} construction \cite{reinohl,fahs}, the nominal \textit{\'{s}aga} `sangha' takes genitive case marking. In contrast, Ashokan Prakrit employs either a dative dependent (e.g. \textit{etāya}) or the stem-compounding strategy described above. It cannot be ruled out, however, that the modern Sinhala and Dhivehi datives originate in a similar compound-like use of \textit{ártʰāya} \cite{fritz}.

} 

\subsubsection{Other cases}

Another unexpected sandhi was observed in Girnar 2:2, \textit{manusōpagāni ca pasōpagāni ca} `beneficial to man and beneficial to animal'. The form \textit{pasōpagāni} is underlying \textit{pasu} `(domestic) animal' + \textit{upagāni} `benefits', wherein the sandhi of \textit{u + u} gives \textit{ō} rather than expected \textit{ū} (as in Sanskrit) or \textit{u} (as in Pali). This sandhi is found in every other edition of the edict; Jaugada even has \textit{pasu-ōpagāni}. Like the previous example, we could claim that \textit{upagāni} is undergoing grammaticalized to a benefactive postposition here, but we feel it is too speculative to claim that, and instead believe it to be phonological analogy with \textit{manusōpagāni}. We analysed it as a noun compound with \ud{DEPREL} \ud{nmod}.

\section{Future work}

The main task ahead of us involves completing annotation, which will require gathering and critical editing of Ashokan texts discovered in the past century that are yet to be digitally compiled. What has been annotated already will be included within the next annual UD corpus release.

After a good selection of annotated inscriptions from several dialects is available, we will make use of computational methods to analyze the corpus. Automatic word-level alignment between dialectal variants of the same edict will enable us to compare dependency structure, case marking, sound change outcomes, along with other dialectal features. On the technical side, we would also like to see if training data from Sanskrit with finetuning on the smaller Ashokan corpus could be used to automatically perform UD annotation of texts in other Middle Indo-Aryan languages.

More broadly, we would like to continue UD annotation of texts in earlier Indo-Aryan languages in order to have data to better address historical linguistic questions. Given the value already demonstrated by corpus data for Indo-Aryan historical linguistics \cite{stronski2020shaping}, open-access corpora annotated using Universal Dependencies, with fine-grained analyses of morphology and syntax beyond individual glossed examples, will surely help put some of the controversial issues in the field to rest. Comparisons of Ashokan Prakrit with other stages of Indo-Aryan will help us study language change, e.g.~the development of configurationality in Middle Indo-Aryan \cite{reinohl}. Dialectal variation (and possible substrate influence) in Ashokan Prakrit should also be studied in comparison with regional NIA data. Other recent work in computational approaches to this area \cite{cathcart-rama-2020-disentangling,cathcart2020probabilistic,arora-etal-2021-bhasacitra,jambu} encouraged us to pursue the study of South Asian historical linguistics from a similar angle. 

Some texts we hope to treebank in the future include: the Pāli Canon, plays in the various later Dramatic Prakrits (e.g.~\textit{Gāhā Sattasaī}), the \textit{L\={o}m\={a}f\={a}nu} documents (Old Dhivehi), the \textit{Bāṇāsurakatʰā} (Old Kashmiri), the \textit{Gurū Grantʰ Sāhib} (Sant Bhāṣā, Old Punjabi), the \textit{Caryāpada} (Old Bengali), the \textit{Šāh jō Risālō} (Sindhi), and epics and poetry from the Hindi Belt and Maharashtra. Serious work on typology in South Asia will also require treebanking for Dravidian (which has a long historical attestation), Munda, and other language families of the region.

\section*{Acknowledgements}

We thank John Peterson for his continuous guidance in our work on Ashokan Prakrit, and Amir Zeldes, Nathan Schneider, and Samopriya Basu for their helpful comments on the manuscript. We also thank the three anonymous UDW reviewers for their constructive feedback.



\bibliography{anthology,coling2020}

\end{document}